\documentclass[10pt,twocolumn]{IEEEtran}
\usepackage{graphicx}          
\usepackage{amsmath}
\usepackage{tabularx}
\usepackage{amsfonts}
\usepackage{url}
\usepackage{verbatim}
\usepackage{times,epsfig}
\usepackage{makecell}
\usepackage{psfrag}
\usepackage{subfigure}
\usepackage{stfloats}
\usepackage{setspace}
\usepackage{amssymb}
\usepackage{multirow}
\usepackage{colortbl}
\usepackage[justification=centering]{caption}
\usepackage{fancyhdr}
\usepackage[table,xcdraw]{xcolor}

\begin{document}
	
	\pagenumbering{arabic}
	
	\title{When Quantum Information Technologies Meet Blockchain in Web 3.0}
	\author{Minrui Xu, Xiaoxu Ren, Dusit Niyato,~\emph{Fellow, IEEE}, Jiawen Kang,~\emph{Member, IEEE}, Chao Qiu~\emph{Member, IEEE}, \\Zehui Xiong,~\emph{Member, IEEE}, Xiaofei Wang,~\emph{Senior Member, IEEE}, and Victor C. M. Leung,~\emph{Life Fellow, IEEE}
	\thanks{M.~Xu and D.~Niyato are with the School of Computer Science and Engineering, Nanyang Technological University, Singapore (e-mail: minrui001@e.ntu.edu.sg; dniyato@ntu.edu.sg). X. Wang, X. Ren, C. Qiu are with the College of Intelligence and Computing, Tianjin University, Tianjin 300072, China. X. Wang and C. Qiu are also with the Guangdong Laboratory of Artificial Intelligence and Digital Economy (SZ), Shenzhen 518000, China (e-mails: xiaofeiwang@tju.edu.cn, xiaoxuren@tju.edu.cn, chao.qiu@tju.edu.cn). J.~Kang is with the School of Automation, Guangdong University of Technology, China (e-mail: kavinkang@gdut.edu.cn). Z.~Xiong is with the Pillar of Information Systems Technology and Design, Singapore University of Technology and Design, Singapore 487372, Singapore (e-mail: zehui\_xiong@sutd.edu.sg). 
 Victor~C.M.~Leung is with the College of Computer Science and Software Engineering, Shenzhen University, Shenzhen 518061, China, and also with the Department of Electrical and Computer Engineering, The University of British Columbia, Vancouver BC V6T 1Z4, Canada (E-mail: vleung@ieee.org).
 }
	}
	\maketitle
	\pagestyle{headings}

	\begin{abstract}
		With the drive to create a decentralized digital economy, Web 3.0 has become a cornerstone of digital transformation, developed on the basis of computing-force networking, distributed data storage, and blockchain. With the rapid realization of quantum devices, Web 3.0 is being developed in parallel with the deployment of quantum cloud computing and quantum Internet. In this regard, quantum computing first disrupts the original cryptographic systems that protect data security while reshaping modern cryptography with the advantages of quantum computing and communication. Therefore, in this paper, we introduce a quantum blockchain-driven Web 3.0 framework that provides information-theoretic security for decentralized data transferring and payment transactions. First, we present the framework of quantum blockchain-driven Web 3.0 with future-proof security during the transmission of data and transaction information. Next, we discuss the potential applications and challenges of implementing quantum blockchain in Web 3.0. Finally, we describe a use case for quantum non-fungible tokens (NFTs) and propose a quantum deep learning-based optimal auction for NFT trading to maximize the achievable revenue for sufficient liquidity in Web 3.0. In this way, the proposed framework can achieve proven security and sustainability for the next-generation decentralized digital society.
	\end{abstract}

	
\begin{figure*}[!t]

\centering

\includegraphics[width=18cm]{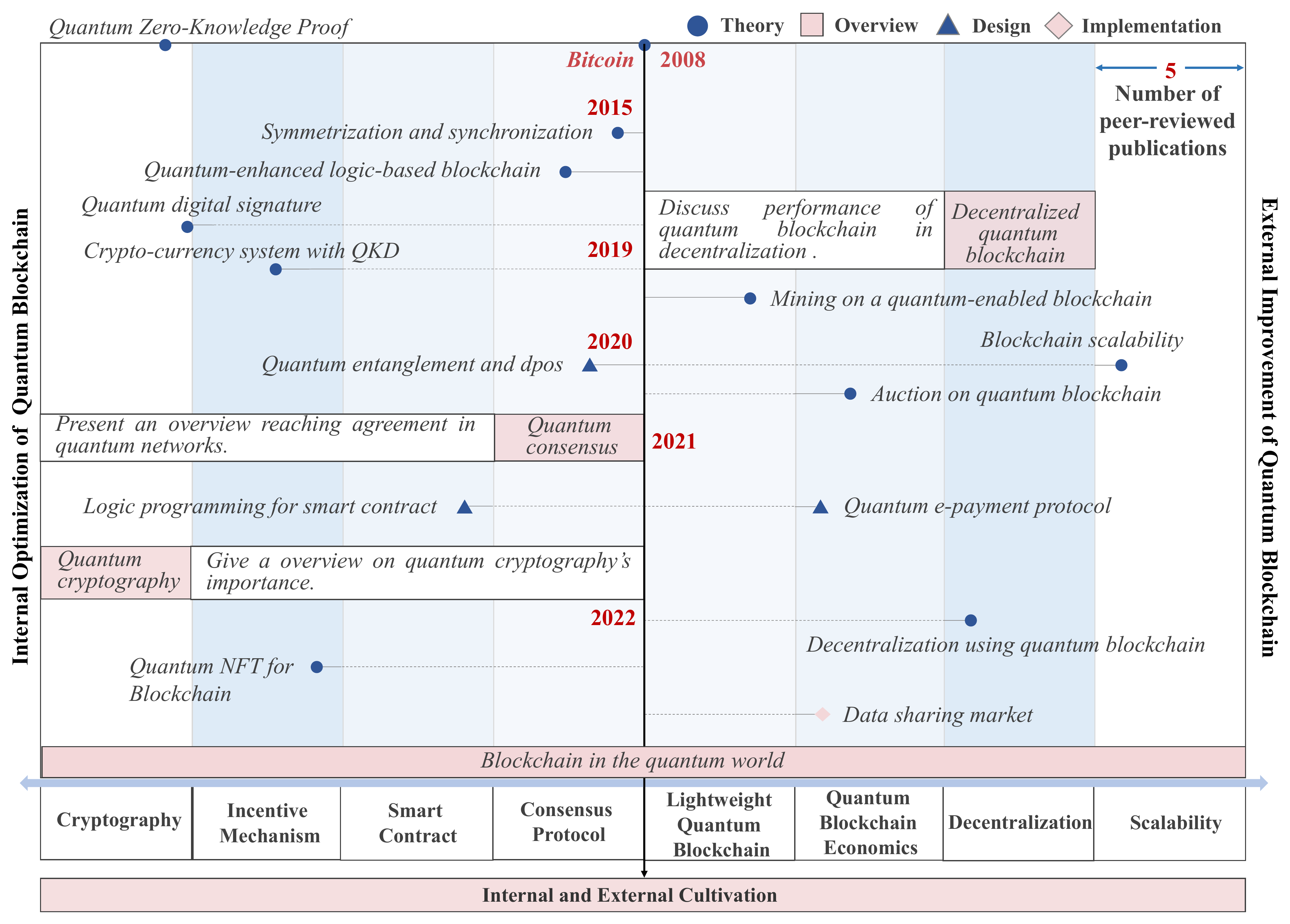}

\caption{An illustration of quantum blockchain-related research activities. }

\label{fig:roadmap}

\end{figure*} 	
	\section{Introduction}

        With distinctive features of the creator economy and decentralized autonomous organization, Web 3.0 can improve user experience and guarantee user privacy and security to the next dimension~\cite{chen2022digital}. The decentralization of Web 3.0 is based on blockchain, along with computing-force networking (CPN)~\cite{wang2020net} and distributed data storage as key technologies, which promises to provide data security, transparency, and accountability for digital transformation activities~\cite{lin2022unified}. These promising advantages of blockchain are based on one-way mathematical functions in classical computing, e.g., hash functions and public-key encryption. Unfortunately, blockchain is particularly at risk since these computational assumptions of one-way functions are broken by quantum computing. With the superposition and the entanglement of quantum bits (qubits), quantum computing can provide sufficient computing power to break existing digital signatures and hash functions within secrecy periods, e.g., quantum computers with 2$\times$10$^7$ qubits can crack RSA-2048 within eight hours.

        While Web 3.0 enables users to read, write, and own their user-generated content (UGC)~\cite{lin2022unified}, classical blockchain is about to cease once the power of a quantum computer meets the requirements to run Shor's algorithm and Grover's algorithm~\cite{fedorov2018quantum}. The development of quantum information technology has opened another door for cryptography. For example, quantum key distribution (QKD) protocols encode the key used for symmetric encryption into a quantum state for transmission over a quantum channel. Together with one-time-pad technology, QKD-secured communication can achieve information-theoretically secure data transmission. Moreover, quantum blockchain can use the quantum hash function and one-way quantum computing functions to develop quantum voting~\cite{pirandola2020advances} and quantum signature~\cite{wang2022quantum} algorithms. In this way, the classical blockchain is updated to the quantum blockchain with the advantage of longer data encryption and utilization of more than twenty years.

        Based on quantum cryptography techniques~\cite{pirandola2020advances}, e.g., quantum hash functions and quantum signature algorithms, quantum blockchain provides several advanced encryption services to quantum-driven Web 3.0. For example, quantum decentralized digital identity (DDID) is fundamental of quantum-driven Web 3.0 that allows users to access decentralized financial (Defi) and decentralized applications (dApps) with the corresponding quantum digital signature. In addition, quantum non-fungible tokens (NFTs)~\cite{pandey2022efficient} represented by quantum hypergraph status are designed to provide a reliable and cheaper digital asset identification mechanism than the classical one. Finally, quantum consensus algorithms, such as quantum delegated proof of stake constructed by quantum voting algorithms, can help to build trust among multiple quantum participants without an authenticated third party~\cite{li2022efficient}.

        In addition to the proven security provided by quantum cryptography, incentive mechanisms in blockchain~\cite{luong2018optimal} should be well-designed to ensure the active participation of miners and NFT buyers for the sustainability of Web 3.0. Especially, sufficient liquidity relies on the achievable revenue from NFT trading market of Web 3.0. However, it is challenging to achieve revenue maximization in the NFT trading market while ensuring NFT buyers' individual rationality (IR) and dominant strategy-incentive compatibility (DSIC). IR indicates all the buyers receive non-negative incentives, while DSIC means no buyers can obtain a higher utility by submitting an untruthful bid. In literature, several deep learning (DL)-based optimal auctions validated the feasibility of incentivizing miners in blockchain~\cite{luong2018optimal}. Therefore, in this paper, we propose a quantum deep learning (QDL)-based optimal auction based on quantum neural networks to achieve optimal auction in NFT trading. The experimental results demonstrate the effectiveness of the proposed QDL-based optimal auction.







The main contributions of this article are summarized as follows:
 \begin{enumerate}
     \item We propose a novel quantum blockchain framework to drive Web 3.0 with information-theoretically security. In the framework, quantum cryptography is leveraged to provide the potential to encrypt and protect the decentralized data of users longer than secrecy periods.
     \item Based on quantum blockchain-based services in the framework, we discuss the potential applications of digital transformation and the corresponding challenges. Furthermore, we identify the benefits of the framework to facilitate the creation, management, and usage of these applications.
     \item To maintain sufficient liquidity in the framework, we propose a quantum machine learning-based optimal incentive mechanism to maximize the achievable revenue in the NFT trading market while guaranteeing buyers' IR and DSIC. In this way, the incentive mechanism can provide long-term sustainability with proven security for the digital society.
 \end{enumerate}
 In literature, several quantum-safe blockchain solutions are proposed, the major ones of which are listed in Fig.~\ref{fig:roadmap}. There are three main types of existing quantum blockchains. The first type uses post-quantum encryption algorithms to protect against attacks from quantum computers. Second, quantum networks have absolute security properties that can be used to guarantee the proven security of quantum blockchain running on classical computers during data transmission. Finally, quantum blockchain running on quantum computers ensures the secure operation of blockchain in quantum computing by using advanced quantum computing techniques to implement quantum signatures and quantum hash functions.

  \begin{figure*}[!t]

\centering

\includegraphics[width=18cm]{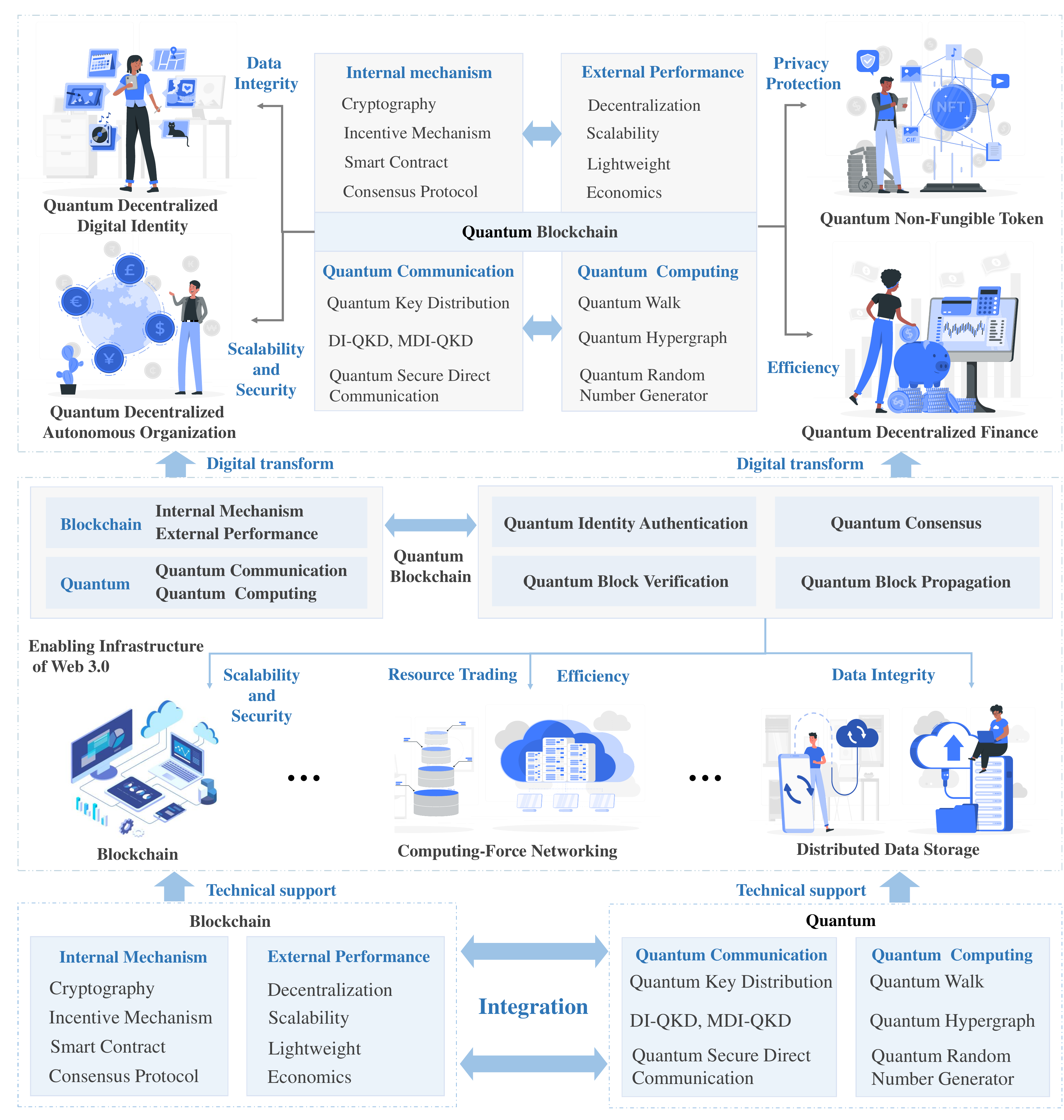}

\caption{The framework of quantum blockchain-driven Web 3.0}

\label{fig:roadmap}

\end{figure*}

\section{The Framework of Quantum Blockchain-driven Web 3.0}
To support the decentralized digital society in the quantum era, the framework of quantum blockchain-driven Web 3.0 consists of enabling infrastructure, quantum cryptography protocols, and quantum blockchain-based services.
\subsection{Enabling Infrastructure of Web 3.0}
In Web 3.0, users can fully control their own digital identities and digital assets based on the critical infrastructure of Web 3.0, including CPN, distribution data storage, and blockchain. Ubiquitous network connectivity enables computing and storage resources in different geographical locations to be merged into a common CPN to support Web 3.0 applications that are connected to the blockchain for computational security. By merging computing resources from end devices, edge servers, and the cloud (including classical and quantum cloud computing) in CPN, Web 3.0 applications can be used for record-keeping, proof of ownership, scheduling, and trading of decentralized digital finance and marketplaces. Distributed storage enables decentralized applications to store large amounts of data and digital assets in distributed storage nodes while recording the data retrieval links used for data identification. Finally, dApps developed based on blockchain act as distributed ledgers for digital identities and digital assets, which provide unified access for Web 3.0 players.

\subsection{Quantum Cryptography Protocols for Web 3.0}

In Web 3.0, several cryptography protocols are adopted to ensure the security and privacy of users, including identity authentication, consensus mechanism, block verification, and block propagation.

\subsubsection{Quantum Identity Authentication} Quantum identity authentication (QIA) uses quantum cryptography protocols to verify the identity of blockchain participants and prevent adversaries from impersonating blockchain participants. To implement information-theoretically secure authentication in quantum blockchain, QIA protocols can use two types of methods for authentication based on shared classical keys and shared entangled states. In QIA protocols with shared classical keys, the two communicating parties share a predefined message in advance to indicate their identity. In QIA protocols with shared entanglement, the communicating parties share a set of entangled particles (i.e., qubits), and each party owns one of each pair of entangled particles and performs the corresponding operation on the entangled pair to identify each other. This method requires plenty of time to store a large number of entangled particles and is not easy to implement. To achieve information-theoretic security, the key or entanglement state that users share in advance in the QIA protocol should ensure that it cannot be obtained by an eavesdropper during use and generally cannot be reused. In addition, identity authentication should be performed concurrently with protocols such as QKD to prevent an eavesdropper from skipping the authentication phase and sharing keys directly.
\subsubsection{Quantum Consensus Mechanism} The consensus mechanism allows nodes in a blockchain network to agree without a trusted third party. In classical blockchain, consensus protocols such as Proof-of-Work (PoW), Proof-of-Stake (PoS), Delegated Proof-of-Stake (DPoS), and practical Byzantine Fault Tolerance (PBFT) are widely used to protect the block generation process of blockchain. In quantum blockchain, quantum cryptographic algorithms, such as quantum voting, can be used to develop quantum consensus mechanisms. Based on quantum voting, quantum delegated proof-of-stake (QDPoS) is used to defend against quantum attacks on the selection of representative nodes participating in the consensus. Meanwhile, DPoS with node behavior and Borda count (DPoSB)~\cite{wang2022quantum} is proposed to select fairly the witness nodes with low energy consumption.
 \subsubsection{Quantum Block Verification} Before a new block is added to the blockchain, block verification must be performed to check the transaction messages and other related information in the new block. The quantum blockchain generates verifiable information through quantum signature algorithms and sends it to other nodes for verification. After the new block is verified, the quantum miner sends the new block to the other nodes in the blockchain network for consensus. In the quantum blockchain protocol proposed in~\cite{qu2022quantum}, each block is represented by a qubit encoding the information of weighted hypergraph status, which decreases the required storage space of the whole blockchain. This protocol can withstand several security threats, including external attacks, intercept-measure-repeat attacks, and entanglement-measure attacks.
 \subsubsection{Quantum Block Propagation} In the quantum blockchain, the propagation of blocks can be conducted on both classical and quantum networks via quantum secure communication~\cite{coladangelo2020quantum}. When a channel in a network uses QKD to secure its communication, a quantum security key to encrypt the block must be distributed over the quantum channel before block propagation. In addition, quantum blockchain can propagate the qubit used to represent the blockchain information through the entanglements in the quantum network. Both approaches can provide information-theoretic protection for block propagation in the quantum blockchain. However, as quantum communication resources remain scarce and expensive, it is necessary to clarify how to balance the tradeoff of block propagation efficiency and security of blockchain with minimized network costs.

\subsection{Quantum Blockchain-based Services in Web 3.0}
With quantum cryptography protocols, a few promising services based on quantum blockchain are provisioned in Web 3.0, including quantum DDID, quantum NFT (QNFT), quantum DeFi, and quantum DAO.
 \subsubsection{Quantum Decentralized Digital Identity} Users in Web 3.0 prove their ownership of digital assets (e.g., UGC, browsing records, transaction records, and behavior records) in the blockchain by digitally signing with their private keys, i.e., distributed digital identities. Based on the randomness of quantum physics, quantum signature algorithms provide tamper-proof, anonymous, and transparent digital identities for users in Web 3.0. Moreover, quantum DDID can be leveraged to restore the original value in users' accounts~\cite{coladangelo2020quantum}.
\subsubsection{Quantum Non-fungible Token} Non-fungible Token is a unique implementation of Web 3.0 for enforcing user-created content. By tagging digital assets with specific rights, NFT based on smart contracts can generate a unique link, which is recorded on the corresponding blockchain with which cannot be tampered, creating a unique digital collection that can prove ownership. Instead of physically distributing the NFT directly to the owner, quantum algorithms, e.g., quantum walk, can represent the information in the QNFT with quantum bipartite hypergraph states and attach the quantum status to the blockchain. As a result, quantum blockchain can provide more reliable and cost-effective QNFTs with the guarantee of uniqueness and proof of ownership.
 \subsubsection{Quantum Decentralized Finance} Decentralized Finance (DeFi) in Web 3.0, including stablecoins, decentralized exchanges, and peer-to-peer lending, can increase the liquidity of digital assets which are circulating on the blockchain. Quantum blockchain can provide information-theoretically security for transactions in DeFi with instantaneous payment transmission. For example, a hybrid classical-quantum framework is proposed in~\cite{coladangelo2020quantum} to provide a scalable solution to the existing smart contract-based blockchain. In detail, quantum lightning based on quantum secure communication is leveraged to enable quantum DeFi in Web 3.0 with unbounded throughput and instantaneous payments.
 \subsubsection{Quantum Decentralized Autonomous Organization} A decentralized autonomous organization is a form of social organization on Web 3.0 that fully participates in decision-making in a low-trust model. Participants in a DAO have the equal right to determine the rules and processes of collaboration in the blockchain openly and transparently through voting, including smart contracts, prior commitment, task assignment, payroll, and organizational governance. In DAO, the identity of participating users can be verified through the quantum signature algorithm. Then, through the quantum voting mechanism, users can participate in voting on new proposals at DAO in the form of information-theoretical security through the mechanism of quantum voting. Finally, incentives in DAO can be distributed through tokens of quantum blockchain~\cite{coladangelo2020quantum}.

In summary, the proposed framework quantum blockchain-driven Web 3.0 can provide proven security for quantum blockchain-based services by leveraging quantum cryptography protocols. To support the sustainable use of quantum cryptography techniques, efficient networking for blockchain via quantum networks and incentive mechanisms for miners and NFT traders with quantum resources should be achieved.
\begin{figure}
    \centering
    \includegraphics[width=1\linewidth]{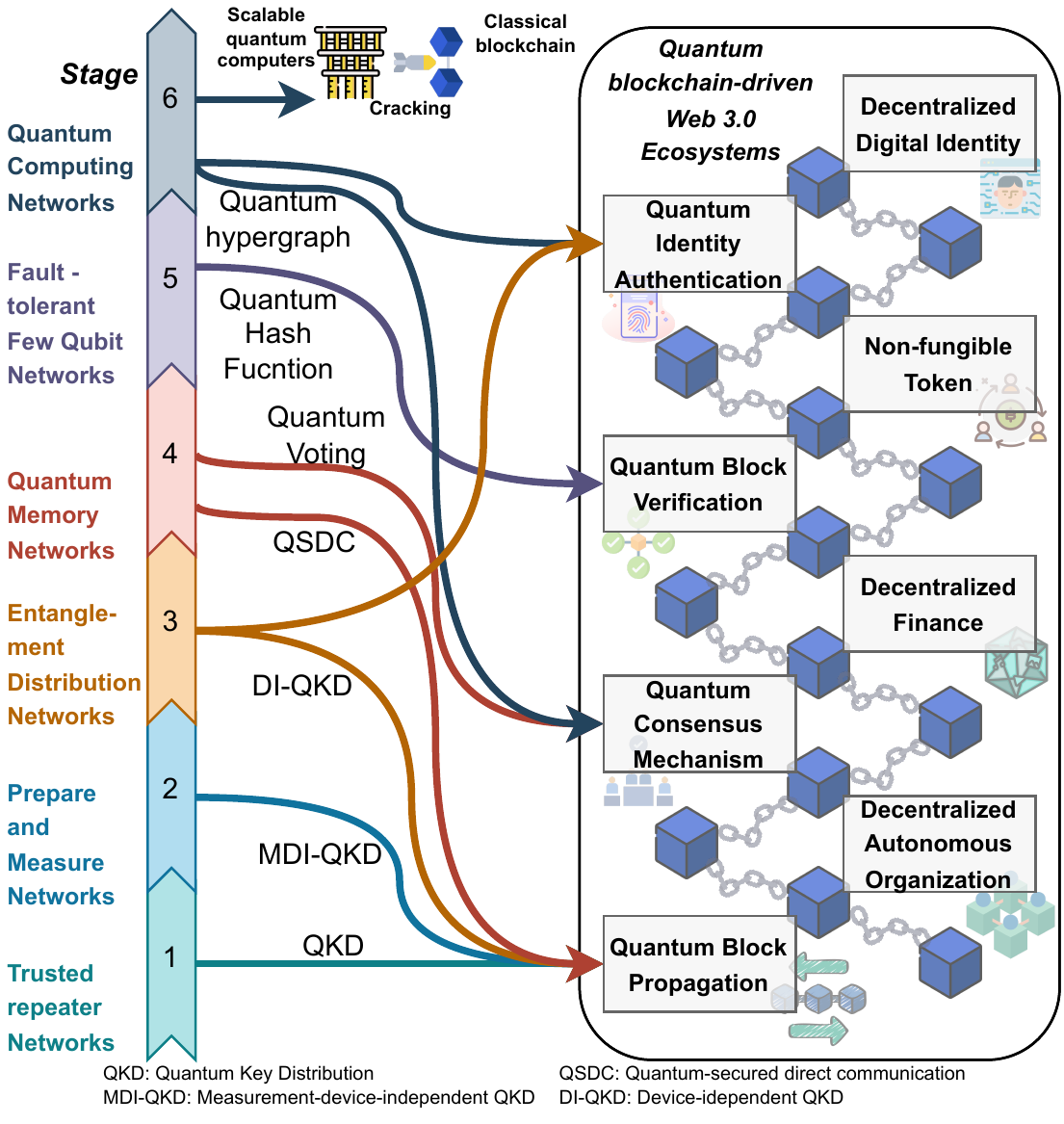}
    \caption{Stages in the development of quantum blockchain-driven Web 3.0. }
    \label{fig:stages}
\end{figure}
\section{Techniques, Applications, and Challenges of Quantum blockchain-driven Web 3.0}
 \subsection{Quantum Cryptography Techniques}
 There are six main stages in the continuous development of quantum blockchain-driven Web 3.0, as shown in Fig.~\ref{fig:stages}. At different stages, quantum cryptography techniques will become mainstream to support the quantum cryptography protocols in that stage.
 \subsubsection{Quantum Secure Communication}
 To achieve communication security in quantum blockchain, QKD and quantum secure direct communication (QSDC) can be adopted. 
 On the one hand, QKD is a key exchange protocol for symmetric encryption via quantum communication. Using the non-cloning theorem and Heisenberg's uncertainty theorem, QKD can provide provable security for communication in the case of the one-time-pad (OTP). In the stage of trusted repeater networks, the vanilla protocols, such as the BB84 protocol~\cite{cao2022evolution}, are proposed to encode key information into quantum states and transmit it via experimental quantum networks. Then, in the stage of prepare-and-measure networks, measurement-device-independent (MDI)-QKD protocols are proposed to improve the feasible distance of QKD links. Finally, in the stage of entanglement distribution networks, device-independent (DI)-QKD protocols are developed to provide a longer entanglement distance of quantum networks. On the other hand, with the development of quantum routers with quantum memory, QSDC becomes possible in the stage of quantum memory networks. The QSDC protocol achieves direct transmission of secret information through chunking, quantum privacy amplification, and error correction. In this way, QSDC constitutes not only a new secure communication paradigm but also an instantaneous communication scheme by transmitting messages in the quantum channel.
 





 
 \subsubsection{Quantum Random Number Generators} Based on the uncertainty principle of quantum mechanics, quantum computing can provide true random number generators for cryptography protocols, i.e., quantum random number generators (QRNG), compared to pseudo-random number generators in classical computing. The advantages of QRNG are multiple. These include the fundamental benefits of exploiting quantum uncertainty, often through the use of photonics for faster performance. Finally, the ability to understand and verify the sources of unpredictability is a core guarantee for quantum blockchain.
 \subsubsection{Quantum Signature Algorithms} Since classical digital signatures based on asymmetric encryption, such as RSA, will be cracked by Shor's algorithm, quantum digital signature algorithms are considered to be an important component of quantum blockchain to ensure tamper-resistance. Based on the computational problems of distinguishability of quantum states, quantum signature algorithms can secure payment and data transferring information in quantum blockchain.
 \subsubsection{Quantum Hash Functions} In the block header of the quantum blockchain, quantum walking can provide a hash function for block verification. Quantum computers can be used not only to crack classical hash functions but also to develop quantum hash functions using quantum one-way functions or quantum random processes. For example, in the quantum blockchain protocol proposed in~\cite{qu2022quantum}, the hash value of each block is encoded in a qubit based on a controlled alternative quantum walk method for generating the hash value of a block.
 \subsubsection{Quantum Voting} Quantum voting protocols~\cite{pirandola2020advances} achieve information-theoretic security of the voting process in the quantum blockchain by sharing quantum information with the requirements of correctness, traceability, verifiability, and anonymity. At the voting setup node, the voting information is generated and encoded with quantum states. Then the voter shares the quantum information with other participants through the quantum channel. At the voting stage, the results of voting with quantum states are calculated in a transparent and anonymous manner.
 \subsection{Potential Digital Transformation Applications}

 \subsubsection{Smart City} Smart cities aim to develop protocols and technologies to improve the quality of People's daily life. Meanwhile, the benefits of smart cities come with other potential risks. For example, massive IoT devices are vulnerable to cyber-attacks. 
Blockchain is a promising solution that enables decentralized and secure data management. 
However, the implementation of quantum computers makes most current encryption algorithms underlying the blockchain open to be hacked. In this regard, quantum cryptography helps to eliminate the risk of data storage and transmission associated with blockchain and IoT.
The work in \cite{ABDELLATIF2021102549} designs a novel authentication and encryption protocol using quantum walks (QIQW). The QIQW protocol enables IoT nodes in smart cities to share data securely and have full control of their records. Meanwhile,  the novel blockchain framework based on QIQW is able to resist probable quantum attacks. 

 \subsubsection{Smart Healthcare} Smart healthcare aims to provide patient-centric healthcare services by secure data collection, efficient data processing, and systematic knowledge extraction \cite{BHAVIN2021102673}. However, maintaining the security and privacy of stakeholders is a challenge for traditional healthcare systems. To improve the efficiency of today’s healthcare system, blockchain emerges as a technology, and it also helps to maintain the security and privacy of all the stakeholders \cite{BHAVIN2021102673}. Replacing traditional encryption signature algorithms with quantum authentication systems, a quantum electronic medical record protocol is designed in~\cite{qu2022quantum}. This protocol tracks every medical record while guaranteeing the security and privacy of EMRs in medical systems. 
\subsubsection{Metaverse} 
In the Metaverse~\cite{xu2022full}, users telepresent as avatars to immerse in and interact with virtual objects in 3D virtual space. The quantum blockchain-driven Web 3.0 is the economic system of Metaverse. For example, avatar identity management can leverage quantum identity authentication for cost-effective verification with proven security. In addition, digital assets in immersive games are minted as QNFTs, which are temper-proof, unique, and tradable. Finally, cross-chain mechanisms in quantum blockchain-driven Web 3.0 can provide instantaneous and information-theoretically secure interoperability for data and payment transferring among different Metaverse.

 %

 \subsection{Main Challenges}
 In the proposed framework, the main challenges can be identified from two aspects: networking and incentives.

 \subsubsection{Quantum Networking for Blockchain in Web 3.0}

 As an inherent property of quantum networking, information-theoretic security can be equipped for networking in Web 3.0. Therefore, quantum networking for blockchain in Web 3.0 is a promising solution. However, scalability in blockchain is still a challenge for consensus and recording payment information in Web 3.0. Currently, three mainstream schemes are proposed to improve throughput for classical blockchain, i.e., payment channels, sharding, and cross-chain. First, with the hash lock in smart contracts, two nodes in quantum blockchain can open a payment channel over a quantum channel. Second, quantum network resources can be leveraged for inter-shard communication for real-time synchronization among multiple shards in a blockchain. Finally, quantum information technologies can help to select the relays for cross-chain transactions among multiple blockchains. However, quantum network resources are limited and insufficient to satisfy the required throughput for blockchain scalability. 

 \subsubsection{Incentive Mechanism Design}

 In Web 3.0, sufficient incentives are required to ensure liquidity in the economic ecosystem for long-term sustainability. On the one hand, the mechanism must provide proper incentives and penalties for achieving consensus in  quantum blockchain based on the devoted quantum computing and communication resources of miners. On the other hand, the mechanism should provide sufficient revenues for NFT trading in Web 3.0. In this way, the proposed framework is sustainable.

\begin{figure}
    \centering
    \includegraphics[width=1\linewidth]{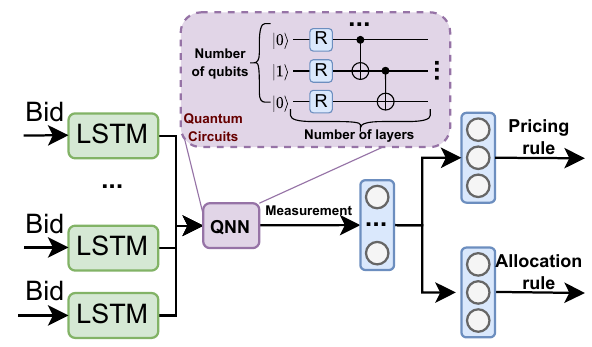}
    \caption{An illustration of the quantum deep learning-based auction with hybrid classic-quantum neural networks.}
    \label{fig:QDL}
\end{figure}
\section{Sustainable Incentive Mechanism for NFT Trading in Quantum Blockchain-driven Web 3.0}

\subsection{The Quantum NFT Protocol}
In quantum blockchain, miners choose the winning node that desires to create a QNFT using QDPoS. The winner of the QDPoS~\cite{li2022efficient} creates a quantum state to represent the QNFT. In the quantum state of the QNFT, the first qubit is encoded with the information (e.g., the name of the UGC) about the UGC, and the second qubit is encoded with the information of the UGC, which is a random phase according to the consensus. After the quantum state of the QNFT is prepared, the creator of the QNFT sends a copy of its state to every participant in quantum blockchain. After receiving the proposed QNFT, each node verifies the QNFT by using quantum gates for each qubit of the QNFT. If the proposed QNFT fails to pass, i.e., if the measurement result does not match the prepared result, the protocol aborts this QNFT and penalizes the proposed peer. If the proposed state passes verification, each node adds the block to its local copy of the blockchain. In this way, the QNFT is successfully added to the quantum blockchain. 
\subsection{Quantum Deep Learning-based Optimal Auction}
To maintain the sustainability of Web 3.0, NFTs can be traded in the market for distribution even after it has been created. The profits generated by NFT transactions can be used for liquidity in the Web 3.0 economic system. The goal of the optimal auction with multiple buyers and multiple items is to maximize the expected revenue while ensuring individual rationality and DSIC~\cite{luong2018optimal}. In the classical deep learning-based optimal auction (DLA), the auctioneer's price rule and allocation rule are output by a neural network. To obtain bids for multiple items, long-short time memory (LSTM) nodes are used to extract useful information from the bids. After the hidden layer of the neural network, the ReLu function and the softmax function of the latent output layer are then processed separately to output the price factors and the allocation probability. To reduce the number of parameters in the DLA neural networks, we propose the QDL-based optimal auction (QDLA) by replacing the hidden layer with quantum circuits. The framework of QDLA is shown in Fig.~\ref{fig:QDL}, where the quantum circuits consist of AngleEmbedding and BasicEntangler layers.
\subsection{Experimental Results}
\begin{figure}[t]
\centering
		\subfigure[Revenue v.s. training epochs.]{
			\centering\begin{minipage}[t]{0.7\linewidth}
				\includegraphics[width=1\linewidth]{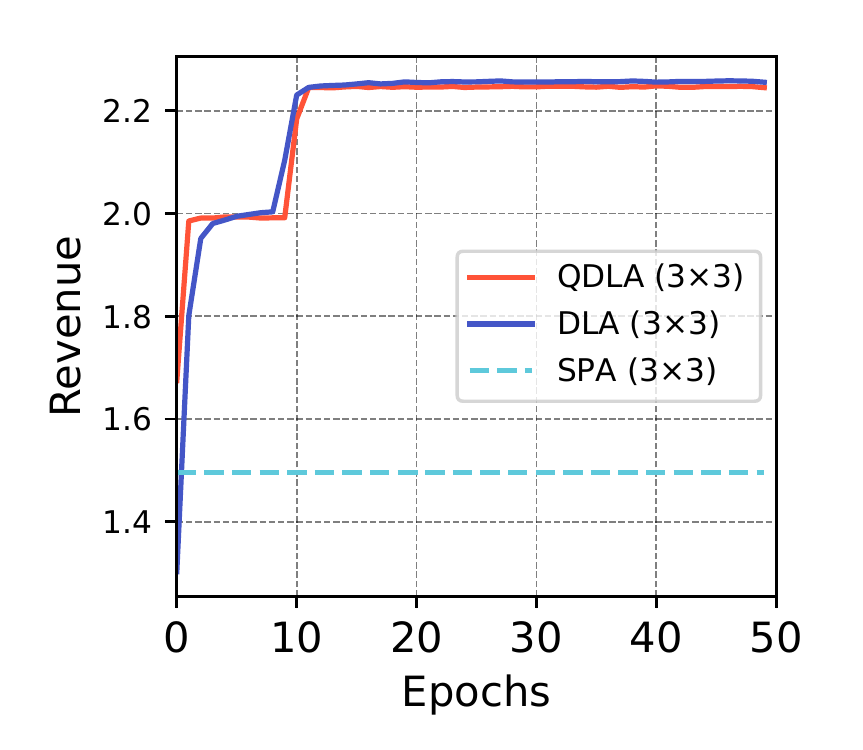}
			\end{minipage}%
		}%

		\subfigure[Regret v.s. training epochs.]{
			\centering\begin{minipage}[t]{0.7\linewidth}
				\includegraphics[width=1\linewidth]{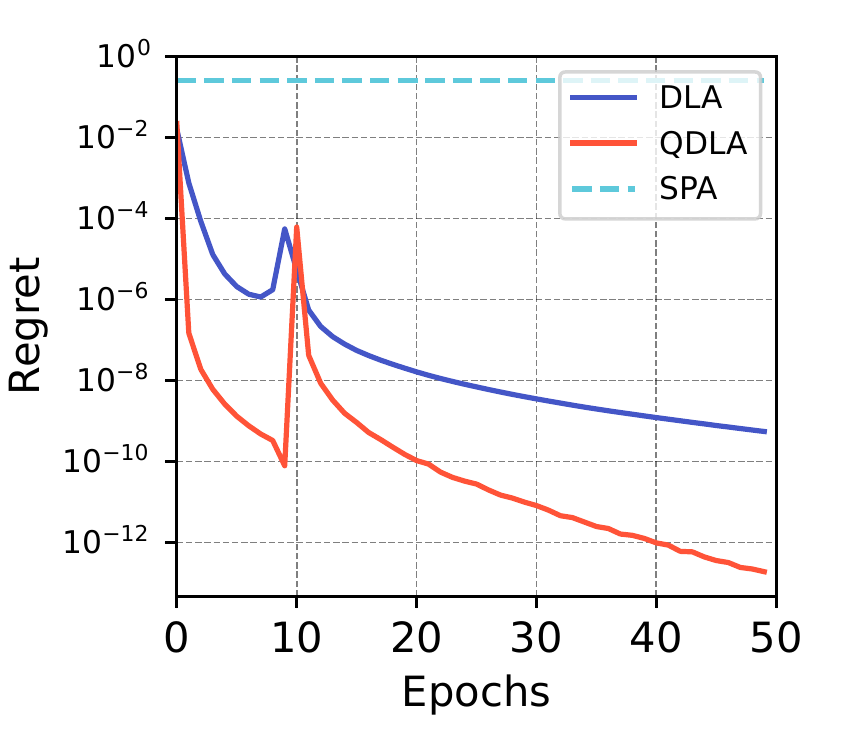}
			\end{minipage}%
		}%
		\caption{The achieved revenue and regret in the NFT trading market.}
		\label{fig:exp}
\end{figure}
To demonstrate the effectiveness of the proposed QDLA, we evaluate the performance of DLA and QDLA in the NFT trading market of Web 3.0. We generate 7000 training samples and 3000 testing samples with 3 buyers and 3 items in the market, and the valuation of NFT is randomly selected from 0 to 1. The learning rate of DLA is set to 0.001, while the learning rate of QDLA is set to 0.01. In DLA, the size of the LSTM layer is 32 and the size of the hidden layer is 32. In classical networks of QDLA, the size of the LSTM layer is 4 and the size of the hidden layer is 16. In quantum networks of QDLA, the number of qubits is 4 and the number of layers is 6. The experimental results of the proposed QDLA are shown in Fig.~\ref{fig:exp}. As we observe in Fig.~\ref{fig:exp}(a), the DLA and the QDLA can converge at about the tenth epoch and outperform the second-price auction (SPA), which can ensure IR and IC but cannot maximize revenue, in terms of expected revenue. The regret, i.e., the square distance between the achieved utility and the optimal utility, is evaluated in Fig.~\ref{fig:exp}(b). It can be observed that both mechanisms can achieve a non-negative regret close to zero, which means that the DLA and the QDLA can achieve DSIC. In summary, although the QDLA consists of fewer parameters than the DLA, the QDLA can achieve similar revenue compared with the DLA and is more robust than the DLA. 

Quantum blockchain effectively synthesizes the security of quantum networks and the efficiency of quantum computing and quantum algorithms. With this use case, we demonstrate clearly that the proposed framework for building a decentralized digital society can achieve not only information-theoretic security but also long-term sustainability in Web 3.0.






 \section{Conclusion}

 In this paper, we have investigated the quantum blockchain-driven Web 3.0 through the implementation of quantum cryptography protocols in blockchain for decentralization, scalability, and security. In detail, we have proposed the framework of quantum blockchain-driven Web 3.0, which consists of enabling infrastructure, quantum cryptography protocols, and quantum blockchain-based services. Moreover, we have discussed potential applications and challenges in the proposed platform. Finally, the QNFT protocol based on QDPoS has been studied as the use case, and a QDLA-based optimal auction has been proposed for improving fluidity in the NFT trading market. The experimental results have demonstrated the effectiveness and efficiency of the proposed QDLA.

	
	\bibliographystyle{ieeetr}
	\bibliography{main}
	\begin{IEEEbiographynophoto}{Minrui Xu} (minrui001@e.ntu.edu.sg)
    received the B.S. degree from Sun Yat-Sen University, Guangzhou, China, in 2021. He is currently working toward the Ph.D. degree in the School of Computer Science and Engineering, Nanyang Technological University, Singapore. His research interests mainly focus on Metaverse, quantum information technologies, deep reinforcement learning, and mechanism design.
    \end{IEEEbiographynophoto}

    \begin{IEEEbiographynophoto}{Xiaoxu Ren}
        [S’20] (xiaoxuren@tju.edu.cn) is currently pursuing the Ph.D. degree at the College of Intelligence and Computing, Tianjin University, Tianjin, China. She received the B.S. degree from the College of Science, Inner Mongolia University of Technology, China, in 2016. Her current research interests include machine learning, computing power networking, and blockchain.
    \end{IEEEbiographynophoto}
    
    \begin{IEEEbiographynophoto}{Dusit Niyato} [M'09, SM'15, F'17] (dniyato@ntu.edu.sg)
        is currently a professor in the School of Computer Science and Engineering, Nanyang Technological University, Singapore. He received the B.Eng. degree from King Mongkuts Institute of Technology Ladkrabang (KMITL), Thailand in 1999 and Ph.D. in electrical and computer engineering from the University of Manitoba, Canada in 2008. His research interests are in the areas of Internet of Things (IoT), machine learning, and incentive mechanism design.
    \end{IEEEbiographynophoto}
    \begin{IEEEbiographynophoto}{Jiawen Kang} [M'18]
        (kavinkang@gdut.edu.cn) received the M.S. degree and the Ph.D. degree from the Guangdong University of Technology, China, in 2015 and 2018, respectively. He is currently a full professor at the Guangdong University of Technology. He was a postdoc at Nanyang Technological University from 2018 to 2021, Singapore. His research interests mainly focus on blockchain, security, and privacy protection in wireless communications and networking.
    \end{IEEEbiographynophoto}
    \begin{IEEEbiographynophoto}{Chao Qiu} [S’15, M’19] (chao.qiu@tju.edu.cn) is currently a lecturer in the School of Computer Science and Technology, College of Intelligence and Computing, Tianjin University. She received the B.S. degree from China Agricultural University in 2013 in communication engineering and the Ph.D. from Beijing University of Posts and Telecommunications in 2019 in information and communication engineering. From September 2017 to September 2018, she visited Carleton University, Ottawa, ON, Canada, as a visiting scholar. Her current research interests include machine learning, computing power networking and blockchain.
    \end{IEEEbiographynophoto}
    \begin{IEEEbiographynophoto}{Zehui Xiong} [M'20]
        (zehui\_xiong@sutd.edu.sg) is an Assistant Professor at Singapore University of Technology and Design. Prior to that, he was a researcher with Alibaba-NTU Joint Research Institute, Singapore. He received the Ph.D. degree in Computer Science and Engineering at Nanyang Technological University, Singapore. He was a visiting scholar with Princeton University and University of Waterloo. His research interests include wireless communications, network games and economics, blockchain, and edge intelligence.
    \end{IEEEbiographynophoto}
    \begin{IEEEbiographynophoto}{Xiaofei Wang}
         [S’06, M’13, SM’18] (xiaofeiwang@tju.edu.cn) is currently a professor at Tianjin University, China. He received master and doctor degrees from Seoul National University in 2006 to 2013, respectively, and was a post-doctoral fellow with The University of British Columbia from 2014 to 2016. He was a recipient of the National Thousand Talents Plan (Youth) of China. In 2017, he received the Fred W. Ellersick Prize from the IEEE Communication Society. His current research interests include social-aware cloud computing, cooperative cell caching, and mobile traffic offloading.
    \end{IEEEbiographynophoto}
    \begin{IEEEbiographynophoto}{Victor C. M. Leung}
        [S’75, M’89, SM’97, F’03, LF’20] (vleung@ieee.org) is a distinguished professor of computer science and software engineering at Shenzhen University. He is also an emeritus professor of electrical and computer engineering and the Director of the Laboratory for Wireless Networks and Mobile Systems at the University of British Columbia (UBC). His research is in the areas of wireless networks and mobile systems. He is a Fellow of the Royal Society of Canada, a Fellow of the Canadian Academy of Engineering, and a Fellow of the Engineering Institute of Canada.
    \end{IEEEbiographynophoto}
\end{document}